\documentclass[letterpaper, 10 pt, conference]{ieeeconf}  

\IEEEoverridecommandlockouts                              

\usepackage{amsmath,amssymb}
\usepackage[]{breqn}

\usepackage[dvipsnames]{xcolor}

\newcommand{\vvector}[1]{\mathbf{#1}}

\newcommand{\defas}[0]{\ensuremath{\stackrel{\mathrm{\scriptscriptstyle{def}}}{=}}}
\newcommand{\nat}[0]{\ensuremath{\mathbb{N}}}
\newcommand{\natpos}[0]{\ensuremath{\mathbb{N} \setminus \{ 0 \}}}
\newcommand{\real}[0]{\ensuremath{\mathbb{R}}}

\newcommand{\intinterval}[2]{\ensuremath{[{#1} \cdots {#2}]}}
\newcommand{\realinterval}[2]{[{#1}, {#2}]}

\newcommand{\sig}[1]{{\small{\textsf{{#1}}}}}

\newcommand{\CommentedText}[1]{}

\usepackage{graphicx}
\usepackage{caption}
\usepackage{subcaption}
\usepackage{multirow}
\usepackage{url} 
\usepackage{booktabs}

\newcommand{\Comment}[1]{}
\newcommand{\projabbv}{BAM}

\newcommand\defeq{\mathrel{\stackrel{\makebox[0pt]{\mbox{\normalfont\scriptsize def}}}{:=}}}

\title{\LARGE \bf
BAM: Box Abstraction Monitors for Real-time OoD Detection\\ in Object Detection
}

\author{Changshun Wu$^{1}$, Weicheng He$^{1}$, 
Chih-Hong Cheng$^{2}$, Xiaowei Huang$^{3}$, and Saddek Bensalem$^{1}$\\ 
\thanks{$^{1}$ University of Grenoble Alps, Grenoble, France.}
\thanks{$^{2}$ Chalmers University of Technology, Gothenburg, Sweden.}
\thanks{$^{3}$ University of Liverpool, Liverpool, UK.}
\thanks{Correspondence to: changshun.wu@univ-grenoble-alpes.fr}
}

\begin{document}

\maketitle
\thispagestyle{empty}
\pagestyle{empty}

\begin{abstract}
Out-of-distribution (OoD) detection techniques for deep neural networks (DNNs) become crucial thanks to their filtering of abnormal inputs, especially when DNNs are used in safety-critical applications and interact with an open and dynamic environment. Nevertheless, integrating OoD detection into state-of-the-art (SOTA) object detection DNNs poses significant challenges, partly due to the complexity introduced by the SOTA OoD construction methods, which require the modification of DNN architecture and the introduction of complex loss functions. This paper proposes a simple, yet surprisingly effective, method that requires neither retraining nor architectural change in object detection DNN, called \underline{B}ox \underline{A}bstraction-based \underline{M}onitors (\projabbv). The novelty of \projabbv~stems from using a finite union of convex box abstractions to capture the learned features of objects for in-distribution (ID) data, and an important observation that features from OoD data are more likely to fall outside of these boxes. The union of convex regions within the feature space allows the formation of non-convex and interpretable decision boundaries, overcoming the limitations of VOS-like detectors without sacrificing real-time performance. Experiments integrating \projabbv~into Faster R-CNN-based object detection DNNs demonstrate a considerably improved performance against SOTA OoD detection techniques.  
\end{abstract}    
\section{Introduction}

Perception systems are critical for an autonomous robot and, among  techniques to implement perception systems, object detection is a fundamental one.  
Despite remarkable advancements in performance, deep neural network (DNN) based object detection systems are not immune to safety concerns~\cite{abrecht2023deep}, particularly in critical applications that may affect human lives and crucial decision-making processes. For instance, as highlighted in~\cite{li2022coda}, 
DNNs frequently exhibit noticeable performance deterioration when dealing with edge cases.
Moreover, additional mechanism is needed to reject, or assign low prediction confidence to,  unknown samples, particularly those out-of-distribution (OoD) samples that are significantly different from the training dataset.

Nevertheless, integrating OoD detection into object detection DNNs while satisfying real-time requirements is known to be challenging. A recent experimental study~\cite{du2022vos} demonstrated that classical OoD detection algorithms such as softmax extensions~\cite{hendrycks2016baseline}, ODIN~\cite{liang2018enhancing} or energy-based techniques~\cite{liu2020energy}, despite giving relatively satisfactory results in image classification tasks, do not yield satisfactory performance when directly applied to object detection. This pessimistic result leads researchers to  consider  
altering of DNN architecture as well as retraining, 
e.g., VOS~\cite{du2022vos} and  EvCenterNet~\cite{nallapareddy2023evcenternet}. Nevertheless, the training of such networks, as demonstrated by EvCenterNet, turns out to be non-trivial and hardly generalizable due to integrating multiple loss functions into one via introducing hyper-parameters. 
Worse still, in practice, the pre-trained models to be monitored may not be permitted to be altered.

\begin{figure}[t]
    \centering
    \includegraphics[width=1.0\columnwidth]{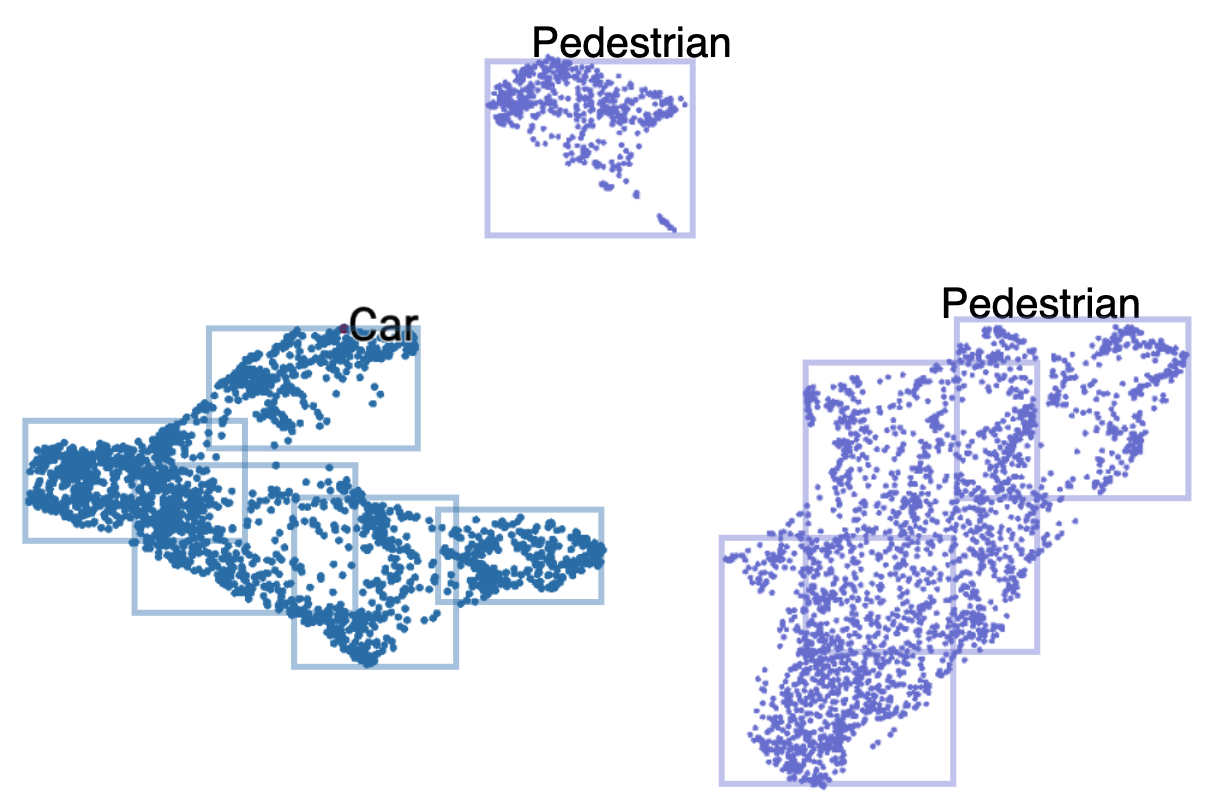}
    \caption{An illustrative example demonstrating the superiority of \projabbv~over the SOTA OoD detection method in object detection, VOS, which assumes a single center of the learned features of each output class and fits a class-conditional Gaussian distribution. However, a well-trained network does not necessarily form a single centered cluster for each output class (cf. the class of pedestrian). Even if it holds, the shape of the cluster does not necessarily have to be a $n$-dimensional ball (cf. the class of car).}
    \label{fig:multiCenters}
    \vspace{-7mm}
\end{figure}

In this paper, we show that such a route via architectural modification and retraining is not really mandatory, and propose \underline{B}ox \underline{A}bstraction-based \underline{M}onitor
for Object Detection, \projabbv~for short, a strikingly direct yet powerful method that empirically demonstrates superior performance compared to the SOTA, without any 
modification to a trained object detection DNN. 
\projabbv\ extends boxed abstraction monitors in classification~\cite{henzinger2020outside,cheng2020towards,cheng2022prioritizing,wu2023customizable} to object detection.   We propose leveraging the finite union of convex polytopes to enclose and characterize the shape of in-distribution (ID) data in the feature space. Our approach is rooted in a pragmatic observation, revealing that networks are not always able to learn a convex decision boundary with a unique center. This is exemplified in Fig.~\ref{fig:multiCenters}, where a natural way of enclosing the decision boundary for ID, as used by BAM, would be to use a union of multiple convex hulls. We also show that compared to existing results in boxed abstraction where data needs to be two-sided, for object detection, building the abstraction with one-sided data suffices to achieve considerable performance gain compared to existing results. Finally, we also detail additional architectural decisions, such as the layer where the monitor is introduced. 

To meet the stringent real-time requirements in object detection, we further consider the shape of the convex polyhedra to be used in the monitors and the number of polyhedra that can be introduced. With the parallelized computation introduced by the GPU, the benefit of using boxes allows highly efficient checking while the memory footprint for the boxes is substantially more compact than dedicated architecture pipelines such as EvCenterNet. 
Although our evaluation is done on Faster R-CNN~\cite{ren2015faster} object detector with a two-stage architecture, we also detail how GPU parallelization enables migrating the techniques into single-stage detectors.
Finally, we extensively evaluate \projabbv~and compare its capabilities against 
sampling-free method VOS~\cite{du2022vos}. On all datasets including KITTI~\cite{Geiger2012CVPR}
and BDD100K~\cite{yu2020bdd100k} datasets, \projabbv~demonstrates its superior performance for the OoD detection, while only introducing~$1.65\%$ overhead compared to the standard Faster R-CNN implementation. 

In summary, our contributions are as follows:
\begin{itemize}
    \item A novel OoD detection framework integrated into two-stage object detectors without the need to modify or fine-tune the detection pipeline.
    %
    %
    \item An extensive evaluation against the SOTA baselines confirming the effectiveness of our methodology. 
     \item A public repository on all our codes, models, and experimental results at \url{https://gricad-gitlab.univ-grenoble-alpes.fr/dnn-safety/bam-ood}.
\end{itemize}

The rest of the paper is organized as follows. After briefly
reviewing related results in Sec.~\ref{sec:relatedWork}, Sec.~\ref{sec:approach} presents details of our \projabbv~approach. Experimental results
are described in Sec.~\ref{sec:experiments}. Finally, the main conclusions and future work are drawn in Sec.~\ref{sec:conclusion}.

\section{Related Work}\label{sec:relatedWork}

Our proposed methodology is part of the active research field regarding OoD, novelty or anomaly detection, and uncertainty quantification (where one can set the proper uncertainty threshold for rejecting an input to be in-distribution), where the review from Salehi~\textit{et~al.}~\cite{salehi2021unified} provides an excellent overview.
However, one of the notable challenges is to migrate the technique into the object detection task while satisfying the real-time constraints by only using limited resources (memories to store the monitor and computing capabilities), which is the primary focus of this paper. 

For OoD detection to be integrated into object detection, one natural method is to consider explicit or implicit ensembles, such as the MC dropout approach. This includes methods such as modifying the detection head to enable dropouts in SSD (Miller~\textit{et~al.}~\cite{miller2018dropout}) or RetinaNet (Harakeh~\textit{et~al.}~\cite{harakeh2020bayesod}), as well as adding an additional dropout layer by extending YOLO (Kraus~\textit{et~al.}~\cite{Kraus_2019}). Nevertheless, the ensemble method requires operating under multiple passes to generate the final decision regarding OoD, thus being computationally more expensive than sampling-free methods similar to our approach. 

For sampling-free methods, we consider the closest work to ours to be VOS~\cite{du2022vos} due to the method also applicable on Faster R-CNN. While the empirical evaluation of our approach demonstrates superiority over VOS, the result can also be justified due to the theoretical analysis that VOS uses a convex shape for characterizing the ID data, while our method is more general than theirs due to the capability to have a finite union of convex polyhedra (boxes) to characterize the decision boundary for ID. The evidence can be observed by the generalizability VOS~\cite{du2022vos}; while it is considered as the SOTA when integrated into Faster R-CNN, its effectiveness diminishes in Transformer-based architectures~\cite{du2022siren}.
The
CertainNet~\cite{gasperini2021certainnet} architecture extends the CenterNet~\cite{duan2019centernet} object detector by learning a set of class representatives called centroids, which are then compared with each prediction at inference time. When considering the decision boundary of centroids, it is again convex, thereby following similar limitations to VOS. Finally, EvCenterNet~\cite{nallapareddy2023evcenternet} also extends CenterNet by integrating evidential learning~\cite{EvidentialClassification}. Both EvCenterNet and CertainNet require modifying architecture while integrating new losses in the training process apart from the standard loss related to improving the prediction quality. Contrarily, our OoD detection approach can be directly integrated into any well-trained object detection network while ensuring real-time performance. 

\section{Technical Approach}\label{sec:approach}

We present our box abstraction-based monitor for Faster R-CNNs in two steps: i) introducing how to construct a box-based monitor for a Faster R-CNN (Sec.~\ref{sec:monitorConstruction}); ii) explaining how to utilize it with the network to identify potential mispredictions in real-time (Sec.~\ref{sec:monitorUtilization}).

\subsection{Basic Notions}
Before introducing the boxed monitor, we review the essential concepts and notation relating to Faster R-CNN networks and the key element in our method, \textit{box abstractions}, an efficient data structure for representing the features corresponding to network detections from ID data. 
Let $\nat$ and~$\real$ be the sets of natural and real numbers.
We use $\intinterval{a}{b}$ with $a, b \in \nat$ and $a \leq b$ to refer to integer intervals.
To refer to real intervals, we use $\realinterval{a}{b}$ with $a, b \in \real \cup \{{-\infty}, \infty\}$ and if $a, b \in \real$, then $a \leq b$. We use a square bracket when both sides are included and a round bracket to exclude endpoints (e.g., $[a, b)$ for excluding~$b$). 
For $n \in \natpos$, $\real^n \defas \underbrace{ \real \times \cdots \times \real}_{n\ \text{times}}$ is the space of real coordinates of dimension~$n$ and its elements are called $n$-dimensional vectors.
We use $\mathbf{x} = (x_1, \ldots, x_n)$ to denote an $n$-dimensional vector and $\mathbf{x}_j$ as the $j$-th element $x_j$ for $j\in [1 \cdots n]$.

\begin{figure*}[t]
   \centering
   \includegraphics[width=\textwidth]{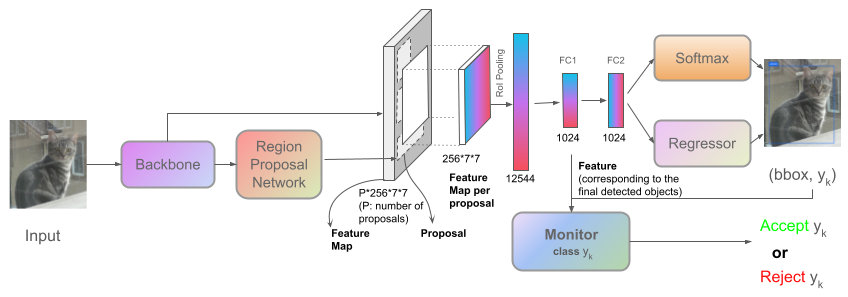}
   \caption{Faster R-CNN architecture and the integration of BAM. For monitor construction, features are extracted from FC1 or the penultimate layer FC2 in the MLP Head of the model. The value~$P$, i.e., the number of proposals per image, equals~$1000$.}
   \label{fig:architecture}
\end{figure*}

\vspace{1mm}
\subsubsection{Faster R-CNN models}
A Faster R-CNN model typically consists of three key components: a backbone network, a region proposal network (RPN), and a region of interest (RoI) head, as shown in Figure~\ref{fig:architecture}.
In general, the model operates according to the following workflow.
The backbone network initially acquires an input image and generates a corresponding feature map.
Subsequently, the RPN utilizes this feature map and anchor information to generate a set of object proposals, each consisting of a tuple of tentative bounding box coordinates and an objectness score indicating the likelihood of containing an object of interest.
Afterward, the RoI head proceeds to process these proposals further, extracting a fixed-length feature vector for each proposal.
Finally, these feature vectors are fed into the Multi-layer perceptron (MLP) head (classifier) for predicting the object category and determining the ultimate bounding box position.

\vspace{1mm}
\subsubsection{Tight box abstraction for a dataset~\cite{henzinger2020outside,cheng2020towards,wu2023customizable}}
In $n$-dimensional geometric space, a \textit{box} is a continuous set, often used to abstract some point sets.
It consists of~$n$ intervals, each corresponding to the upper and lower bounds that each dimension can take.
For a dataset $X=\{\textbf{x}^{1}, \ldots, \textbf{x}^{m} \}$, we define its \textit{tight box abstraction} as $\sig{TBA}(X) \defeq \langle [a_1, b_1], \ldots, [a_n, b_n] \rangle$ as an~$n$ dimensional box, where for $i \in [1 \cdots n]$, $a_i = \mathsf{min}(\{ \textbf{x}^{j}_{i}\})$ and $b_i = \mathsf{max}(\{\textbf{x}^{j}_{i} \})$. 

\vspace{1mm}
\subsubsection{Distance between a data point and a box abstraction}
Given a data point $\vvector{x} = (x_1, \ldots, x_n)$ and a box $B = \langle [a_1, b_1], \ldots, [a_n, b_n] \rangle$, the distance between them, denoted as $d(\vvector{x}, B)$, are defined as the sum over the distances between each $x_i$ and the interval $[a_i, b_i]$, which is defined as follows:
\begin{equation}
    d(x_i, [a_i, b_i]]) = \left\{ \begin{array}{ccl}
    0 & \mbox{if} & a_i \leq x_i \leq b_i \\
    a_i - x_i & \mbox{if} & x_i < a_i \\
    x_i - b_i & \mbox{if} & x_i > b_i \\
    \end{array}\right.
\end{equation}
Intuitively, this distance measures the effort required to move an external data point into a box, under the assumption that each dimension is independent.

\vspace{1mm}
\subsubsection{Enlargement of box abstraction}\label{enlargement}
A tight box for a dataset can be easily enlarged to encompass by using a buffer vector $\delta \defeq [\delta_1, \ldots, \delta_n]$ to relax the lower and upper bounds as follows:
$B_{\delta} = \langle[a_1-\delta_1, b_1+\delta_1], \ldots, [a_n-\delta_n, b_n+\delta_n]\rangle$.

\vspace{1mm}
\subsection{Monitor Construction}\label{sec:monitorConstruction}
\subsubsection{Box-based monitor construction for Faster R-CNNs}
For an object detection network with fixed parameters, let $D_{\text{train}} \defeq \{ (\mathbf{x}, \mathbf{gt})\}$ represent its training dataset $with$ $\mathbf{x}$ being the input and $\mathbf{gt}$ being the associated ground-truth labels. 
For an input $\mathbf{x}$, let $prop_i \in f_{bbrpn}(\mathbf{x})$ be one of the region proposals generated by the backbone and RPN network of the Faster R-CNN, and let  $p_i \defeq (bbox_i, cls_i) = f_{mlp}(prop_i)$ be the output predictions for proposal $prop_i$ passing through the MLP, where $bbox_i$ is predicted bounding box in terms of size, location and orientation, and~$cls_i$ be the \emph{output class} of the predicted bounding box. Let $f^l_{mlp}(prop_i)$ be the feature vector at the $l$-th layer where the monitor is constructed. 
Based on the previous usage of box abstraction~\cite{henzinger2020outside,wu2023customizable}, boxes are generally constructed at fully-connected layers closer to the output, as they provide a more high-level representation of features of input data.

Assume that the faster R-CNN can produce~$Y$ output classes indexed from~$1$ to~$Y$. 
We now describe how to construct a box abstraction monitor $\mathcal{B}^{l}_y = \{ B^{l,1}_y, \ldots, B^{l,k_{y}}_y \}$ at layer~$l$ for each class~$y \in [1 \cdots Y]$ of a Faster R-CNN network using a four-step process:

\begin{itemize}
    \item \textbf{(Step 1: Feature extraction)} For each output class $y \in [1, \ldots, Y]$, Construct $F^l_{y}$ to contain all feature vectors at the $l$-th layer that contribute to a bounding box prediction \footnote{Similar to VOS, we set the confidence score threshold for predictions to maximize the Faster R-CNN model's micro F1 score, thereby preventing low-scoring predictions from being considered as relevant features for monitor construction.} of class~$y$, as formulated in Eq.~\eqref{eq:feature.vector.set}.

\begin{equation}\label{eq:feature.vector.set}
    \begin{split}
        F^l_{y} \defeq \{ f^l_{mlp}(prop_i)\;| \; prop_i \in f_{bbrpn}(\vvector{x}),\;\;\;\;\;\;\;\;\;\;\;\;\\ (\mathbf{x}, \mathbf{gt}) \in D_{\text{train}}, (bbox_i, y) = f_{mlp}(prop_i) \} 
    \end{split}
\end{equation}

Once the feature extraction step is completed, each feature vector is currently implemented in BAM as a one-dimensional tensor, containing~$1024$ neuron activation values for a detected object.
\vspace{1mm}
\item \textbf{(Step 2: Feature vector partitioning)} The set of feature vectors in $F^{l}_{y}$ is partitioned into $k_y$ subsets, denoted as $\pi(F^{l}_{y}) \defeq \{ F^{l,1}_{y}, \ldots, F^{l,k_y}_{y}\}$. The partitioning is done by applying a \textit{k-means}~\cite{lloyd1982least} algorithm to cluster these features: the number of clusters is decided by the \textit{density}, which is the targeted number of data points within each cluster. Given a dataset of size~$m$ and the envisioned density value~$\rho$, the number of clusters to be used in the $k$-means algorithm equals $\lfloor\frac{m}{\rho}\rfloor$. The density is thus a hyper-parameter that decides the number of resulting boxes for a given dataset. To avoid cases when the dataset turns huge, we constraint the maximum $k$ from above with a constant~$T$, reflecting the real-time processing capability of the underlying hardware. In our evaluation, we configure~$T$ to be~$10000$, while~$k$ never exceeds~$8000$. 
\vspace{1mm}
\item \textbf{(Step 3: Abstraction building)} Construct $\mathcal{B}^l_y \defeq \{B^{l,1}_y, \ldots, B^{l,k_y}_y\}$, where for $j \in [1 \cdots k_y]$, $B^{l,j}_y$ is a tight box abstraction for subset $F_{y}^{l,j}$, i.e., $B^{l,j}_y = \sig{TBA}(F_{y}^{l,j})$. 

\vspace{1mm}
\item \textbf{(Step 4: Box enlargement)} The enlargement of boxes is driven by the need to control the true positive rate (TPR) within the in-distribution dataset. To ensure a fair comparison with VOS which uses FPR95 (i.e., false positive rate of OoD samples at a 95\% true positive rate of ID samples) as the evaluation criterion, we also enlarge the box based on FPR95. This is done algorithmically by first sorting all feature vectors (with distance from small to large) that fall outside the boxes created by Step~3. Subsequently, enlarge the box by including sorted feature vectors until TPR95 is reached. Given a feature vector $\vvector{z}$, its distance to the monitor~$\mathcal{B}^l_y$ is defined using Eq.~\eqref{eq:dist.box}, which is the minimum distance to any box in the monitor.  

\begin{equation}\label{eq:dist.box}
    \sig{dist}(\vvector{z}, \mathcal{B}^l_y) \defeq \min_{j \in [1 \cdots k_y]} \{d(\vvector{z}, B^{l,j}_y)\}
\end{equation}

\end{itemize}

\vspace{1mm}
\subsection{Monitor Deployment}\label{sec:monitorUtilization}
Given a neural network $N$ and the boxed abstraction monitor $\{\mathcal{B}^l_1, \ldots, \mathcal{B}^l_{Y}\}$, in runtime, the \textbf{monitor rejects a class~$y$ prediction $p_i \defeq (bbox, y)$ for an input~$\vvector{x}$} generating region proposal~$prop_i$, if $\nexists j \in \intinterval{1}{k_y}: f^{l}(prop_i) \in B^{l,j}_y$.
That is, no box contains the feature vector~$f^{l}(prop_i)$ produced at the~$l$-th layer.
As the containment checking $f^{l}(prop_i) \in B^{l,j}_y$ compares $f^{l}(prop_i)$ against the box's lower and upper bounds on each dimension, it can be done in time \emph{linear to the number of monitored neurons}. 

\vspace{2mm}
In summary, the core idea of BAM is to use a data structure to properly enclose the regions in the feature space where the neural network has made decisions.
When the DNN makes a prediction, BAM evaluates whether the corresponding features for making the prediction fall within the enclosed regions.
If the answer is positive, this decision can be considered similar to some previously observed decision behaviors, which are deemed ID. Otherwise, the decision is diagnosed by BAM as OoD.

\vspace{1mm}
\noindent\textbf{Remark}
While our methodology is illustrated in two-stage detectors, we briefly summarize how to transfer it to the single-stage detector.
Single-stage detectors such as YOLO or SSD maintain a grid of cells, with each cell responsible for predicting if there is an object that is centered in that cell.
A simple (brute-force) method is to create, for each cell, one box abstraction-based monitor. 

\section{Implementation and Experiments}\label{sec:experiments}
In this section, we present the implementation of BAM and experimental results that validate its effectiveness on multiple real-world object detection datasets and network variants.
\subsection{Implementation}
We have implemented BAM using \textit{PyTorch}~\cite{NEURIPS2019_9015}, \textit{Scikit-learn}~\cite{scikit-learn}, the computer vision library \textit{Detectron2}~\cite{wu2019detectron2}, and the object detection dataset management library \textit{Fiftyone}~\cite{moore2020fiftyone}, where in the implementation, we have developed three utility modules for \textit{feature extraction}, \textit{clustering}, and \textit{abstraction building}.
Specifically, the feature extraction module relies on the functionality provided by Detectron2, enabling the loading of the Faster R-CNN model and the generation of predictions on a given dataset.
By default, the Faster R-CNN model can only output the bounding box coordinates for object localization and the classification confidence score.
To facilitate the feature extraction from the intermediate layers of the model, we have implemented a custom forward function extended from Faster R-CNN architecture within Detectron2. 

\subsection{Experiment Setup}
We tested on diverse ID datasets to validate the presented approach, utilizing Faster R-CNN models with different backbones.
To assess the monitor performance of each model variant, we evaluated them against multiple OoD datasets.
In the following, we describe the used datasets, models, the evaluation metric, and the process in detail.

\vspace{1mm}
\subsubsection{Datasets}
We trained our model using two ID object detection datasets specific to the autonomous driving domain: BDD100K~\cite{yu2020bdd100k} and KITTI~\cite{Geiger2012CVPR}.
To evaluate the effectiveness of our monitor in detecting OoD data, we evaluated it on three OoD datasets.
Among these, we use two datasets previously employed in VOS: MS-COCO~\cite{cocodataset} and OpenImages~\cite{Kuznetsova_2020}.
Additionally, we curated an extra OoD dataset derived from PASCAL VOC~\cite{Everingham10}, in which we manually filtered out images containing ID objects.

\vspace{1mm}
\subsubsection{Models}
In this study, our primary focus is the Faster R-CNN object detection architecture.
We trained Faster R-CNN models with a feature pyramid network (FPN) using three backbone architectures: ResNet-50, ResNet-101, and RegNetX-4.0GF.
To attain satisfactory performance, each model variant is fine-tuned for a minimum of 40 epochs, starting from the pre-trained weights on the ImageNet dataset.

\subsubsection{Metrics}
In OoD detection, a ``true positive" refers to an object identified as an ID object that does indeed fall within the ID categories.
In contrast, a ``false positive" means an object identified as an ID object that actually belongs to the OoD categories.
We assess the monitor performance in OoD detection using the standard metric \textit{FPR95}, which represents the false positive rate of OoD samples at a 95\% true positive rate of ID samples.
A lower FPR95 indicates a superior ability of monitor to detect OoD objects while still encompassing 95\% of the true positives from the ID datasets.

\subsubsection{Process}
To evaluate
the developed monitors, performance tests are conducted using OoD datasets.
This assessment allows us to gauge the monitors' capability to accurately identify and reject OoD objects during testing.
At the end of the evaluation process, a confusion matrix is constructed for the monitors, facilitating the computation of FPR95.
These evaluation results are then compared with VOS, which is regarded as the SOTA of OoD detection methods in object detection.

\subsection{Results}
In this section, we present experimental results to demonstrate the effectiveness of our method BAM compared to VOS, which is commonly regarded as the STOA against classical methods such as Softmax~\cite{hendrycks2016baseline}, ODIN~\cite{liang2018enhancing}, Energy~\cite{liu2020energy}.

\vspace{1mm}
\subsubsection{Effectiveness}
Our BAM approach achieves superior performance in terms of FPR95 across datasets.
In Table~\ref{tab:FPR95}, we summarize the comparison between BAM and VOS. In many cases, 14 out of 18, BAM outperforms VOS by a significantly large margin ($>10\%$), without any case being under-performed drastically. Even when we refine the margin to be $5\%$, out of~$18$ OoD cases, BAM is still superior to VOS in~$16$ datasets and on par in~$1$ datasets, with~$1$ being worse. Finally, when comparing performance against absolute values, BAM remains superior to VOS in~$16$ cases. 

\begin{table}[t]
    \centering
    \caption{Comparing BAM (feature vector from FC2Relu) with the state-of-the-art VOS method; $\uparrow$ indicates larger values are better and $\downarrow$ indicates smaller values are better; numbers in \textbf{bold} texts imply superiority of the method, with at least $10\%$ performance increase. Values being underlined imply the performance of two methods being on par.}
    \label{tab:FPR95}
    \resizebox{\columnwidth}{!}{%
        \begin{tabular}{|c|c|c|c|c|}
            \hline
            ID & Backbone  & \begin{tabular}[c]{@{}c@{}} mAP$\uparrow$\\ (ID) \end{tabular} & Method & \begin{tabular}[c]{@{}c@{}}FPR95 $\downarrow$ \\  (OoD: MS-COCO/OpenImage/VOC) \end{tabular} \\ 
            \hline
            \multirow{6}{*}{BDD100k} & 
            \multirow{2}{*}{ResNet50}  & \multirow{2}{*}{31.5} & 
            VOS & 
            48.93 / 41.85 / 53.56\\
            \cline{4-5}  
            &  && BAM & \textbf{37.77} / \textbf{31.25} / \textbf{47.58} \\ 
            \cline{2-5} 

            & \multirow{2}{*}{ResNet101}  & \multirow{2}{*}{32.5} & VOS & 
            37.27 / 21.26 / 45.70\\ 
            \cline{4-5} 
            &  && BAM & \textbf{23.41} / \textbf{8.52} / \textbf{32.47}\\ 
            \cline{2-5} 
            
            & \multirow{2}{*}{RegX4.0}  & \multirow{2}{*}{32.7} & VOS &
            \underline{42.91} / 34.16 / \underline{46.77}\\ 
            \cline{4-5} 
            &  && BAM & \underline{38.87} / \textbf{26.74} / \underline{47.7}\\ 
            \hline
            
            \multirow{6}{*}{KITTI} & 
            \multirow{2}{*}{ResNet50}  & \multirow{2}{*}{79.5} & 
            VOS & 15.97 / 7.51 /19.03\\ 
            \cline{4-5} 
            &  && BAM & \textbf{3.69} / \textbf{1.63} / \textbf{5.24}\\ 
            \cline{2-5}
            
            & \multirow{2}{*}{ResNet101}  & \multirow{2}{*}{86.2} & VOS &
            6.09 / \underline{2.37} / 14.52\\ 
            \cline{4-5} 
            &  && BAM & \textbf{4.24} / \underline{2.54} / \textbf{8.98}\\ 
            \cline{2-5} 
            
            & \multirow{2}{*}{RegX4.0}  & \multirow{2}{*}{79.2} & VOS & 12.52 / \underline{5.46} / 14.87\\
            \cline{4-5} 
            &  && BAM & \textbf{7.48} / \underline{5.07} / \textbf{9.88}\\ 
            \hline
        \end{tabular}%
        }
\end{table}

\begin{figure*}[t]
    \centering
    \includegraphics[width=\textwidth]{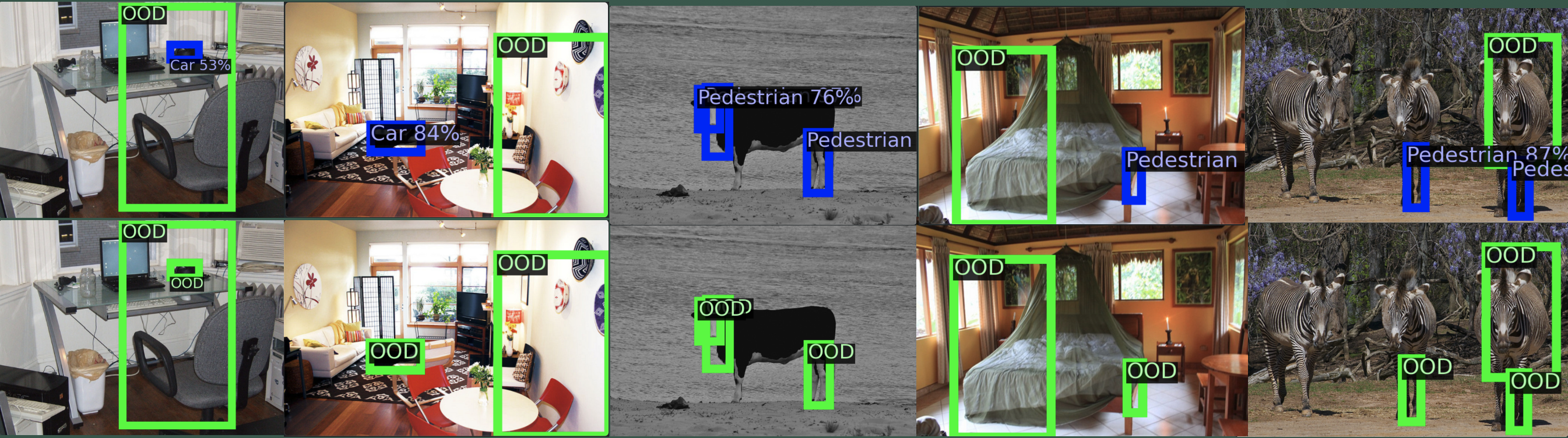}
    \caption{Visualization of detected objects on the OoD images (from MS-COCO) by the benchmark method VOS (top) and BAM (bottom). The in-distribution is KITTI dataset. Blue: Objects detected and classiﬁed as one of the ID classes. Green: OoD objects detected by VOS or BAM, which reduce false positives among detected objects.}
    \label{fig:comparisonVisual}
\end{figure*}

Fig.~\ref{fig:comparisonVisual} presents a qualitative analysis of predictions on several OoD images using object detection models with the benchmark method VOS (top) and BAM (bottom). We use KITTI as the ID dataset. As illustrated by the green bounding boxes in the figure, BAM outperforms the VOS OoD detector in terms of identifying OoD objects, and reducing false positives among detected objects.

\vspace{1mm}
\subsubsection{Execution Time}
On Nvidia RTX A4000 8GB, the GPU-based inference time with BAM averages $41.1$ milliseconds per detection instance on the KITTI dataset, compared to $40.8$ milliseconds without monitoring. This signifies that the monitoring module introduces a negligible additional overhead of~$0.7\%$ ($0.3$ milliseconds).
On the BDD dataset, specifically in the case of the monitor consisting of 7000 boxes for the ``car'' category, the inference time with BAM increases to an average of $94.4$ milliseconds for a single image containing~$1000$ region proposals, as opposed to the original time of $85.4$ milliseconds without monitoring.
Therefore, our BAM approach enhances the inference process without significantly impacting the real-time performance.

\vspace{1mm}
\subsubsection{Ablation study (the impact of layer selection)}
We conducted further investigations to explore the impact of utilizing features from different fully connected layers for monitor construction.
Specifically, we evaluated the performance of the monitors built at layer FC2 and the preceding layer FC1, both with the ReLU activation function.
The representative results are shown in Table~\ref{tab:layers}. Overall, we can observe that the selection of layers can create fluctuation in the performance, as for the KITTI dataset, selecting FC1Relu can achieve better performance. Nevertheless, as demonstrated in Table~\ref{tab:FPR95}, even when selecting the least-performing FC2Relu, the performance is still strictly better than that of VOS.

\begin{table}[h]
\centering
\caption{performance comparison of BAM monitors with features extracted from different layers in the MLP Head. The monitors' performances are consistent across different layers.}
\label{tab:layers}
\resizebox{\columnwidth}{!}{%
\begin{tabular}{|c|c|c|c|}
\hline
ID              & Backbone          & Layer & \begin{tabular}[c]{@{}c@{}}FPR95 \\  (OoD: MS-COCO/OpenImage/VOC-OoD)\end{tabular} \\ \hline
\multirow{4}{*}{KITTI}& \multirow{2}{*}{ResNet50}&        FC1Relu&                                                                                  3.77 / 3.35 / 6.05
\\ \cline{3-4} 
                  &                   &        FC2Relu&                                                                                   \textbf{3.69} / \textbf{1.63} / \textbf{5.24}
\\ \cline{2-4} 
                  & \multirow{2}{*}{RegX4.0}&        FC1Relu&                                                                                 \textbf{7.23} / 7.25 / \textbf{9.01}
\\ \cline{3-4} 
                  &                   &        FC2Relu&                                                                                   7.48 / 
 \textbf{5.07} / 9.88
\\ \hline
\end{tabular}%
}
\end{table}

\vspace{1mm}
\subsubsection{Ablation study (the impact of cluster density)}\label{sub.sub.abloation.density}
\begin{figure}[h!]
    \centering
    \includegraphics[width=\columnwidth]{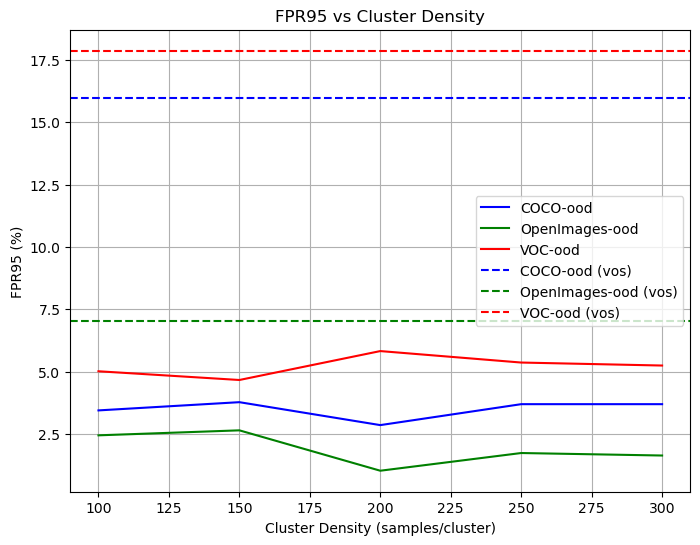}
    \caption{Ablation study on the hyper-parameter $\rho$, density of data points within each cluster. In all settings (varying $\rho$ on x-axis from 100 to 300), our method BAM is better than VOS and performs consistently.}
    \label{fig:ablationK}
\end{figure}

In this ablation study, we vary the hyper-parameter density~$\rho$ to assess its impact on the performance. Remarkably, the results revealed a notable reduction in FPR95 across all density settings, as illustrated in Fig.~\ref{fig:ablationK}. While the study shows that the performance is not sensitive to~$\rho$, the significant decrease in FPR95 across all~$\rho$ values compared to VOS underscores the true effectiveness of the proposed approach.

\section{Conclusion}\label{sec:conclusion}

This paper presented BAM, a box-abstraction-based OoD monitoring method for object detection. BAM nicely enables the characterization of complex and non-convex OoD decision boundaries in the feature space using a finite union of boxes. In addition, BAM can be introduced without the need to change the standard object detection network while maintaining real-time detection capabilities. Our experimental results outperformed the state-of-the-art method by achieving a lower false positive rate of OoD samples while reaching a true positive rate of~$95\%$ for ID samples.
Regarding future research directions, we aim to implement this method in other object detection model families such as YOLO and CenterNets. 
The second direction is to develop a principled approach regarding how to perform refinement when the abstraction is too coarse, as coarse abstraction can negatively influence the decision boundary to manifest false alarms. 
Yet another direction is to consider refining the algorithm by taking into account that a region proposal can have multiple objects. 
Finally, we aim to consider how the construction of monitors can be aligned with safety principles with a clearly specified data quality requirement, such as including a database of edge cases or rare events occurring on the road. This enables an objective evaluation method that is not biased towards specific datasets.









\bibliographystyle{ieeetr}
\bibliography{ref}

\begin{thebibliography}{10}

\bibitem{abrecht2023deep}
S.~Abrecht, A.~Hirsch, S.~Raafatnia, and M.~Woehrle, ``Deep learning safety concerns in automated driving perception,'' {\em arXiv preprint arXiv:2309.03774}, 2023.

\bibitem{li2022coda}
K.~Li, K.~Chen, H.~Wang, L.~Hong, C.~Ye, J.~Han, Y.~Chen, W.~Zhang, C.~Xu, D.-Y. Yeung, {\em et~al.}, ``{CODA}: A real-world road corner case dataset for object detection in autonomous driving,'' in {\em ECCV}, pp.~406--423, Springer, 2022.

\bibitem{du2022vos}
X.~Du, Z.~Wang, M.~Cai, and Y.~Li, ``{VOS}: Learning what you don’t know by virtual outlier synthesis,'' {\em ICLR}, 2022.

\bibitem{hendrycks2016baseline}
D.~Hendrycks and K.~Gimpel, ``A baseline for detecting misclassified and out-of-distribution examples in neural networks,'' in {\em ICLR}, 2016.

\bibitem{liang2018enhancing}
S.~Liang, Y.~Li, and R.~Srikant, ``Enhancing the reliability of out-of-distribution image detection in neural networks,'' in {\em ICLR}, 2018.

\bibitem{liu2020energy}
W.~Liu, X.~Wang, J.~Owens, and Y.~Li, ``Energy-based out-of-distribution detection,'' {\em NeurIPS}, vol.~33, pp.~21464--21475, 2020.

\bibitem{nallapareddy2023evcenternet}
M.~R. Nallapareddy, K.~Sirohi, P.~L. Drews-Jr, W.~Burgard, C.-H. Cheng, and A.~Valada, ``{EvCenterNet}: Uncertainty estimation for object detection using evidential learning,'' in {\em IROS}, pp.~5699--5706, IEEE, 2023.

\bibitem{henzinger2020outside}
T.~A. Henzinger, A.~Lukina, and C.~Schilling, ``Outside the box: Abstraction-based monitoring of neural networks,'' in {\em ECAI}, pp.~2433--2440, IOS Press, 2020.

\bibitem{cheng2020towards}
C.-H. Cheng, C.-H. Huang, T.~Brunner, and V.~Hashemi, ``Towards safety verification of direct perception neural networks,'' in {\em DATE}, pp.~1640--1643, IEEE, 2020.

\bibitem{cheng2022prioritizing}
C.-H. Cheng, C.~Wu, E.~Seferis, and S.~Bensalem, ``Prioritizing corners in ood detectors via symbolic string manipulation,'' in {\em ATVA}, pp.~397--413, Springer, 2022.

\bibitem{wu2023customizable}
C.~Wu, Y.~Falcone, and S.~Bensalem, ``Customizable reference runtime monitoring of neural networks using resolution boxes,'' in {\em RV}, pp.~23--41, Springer, 2023.

\bibitem{ren2015faster}
S.~Ren, K.~He, R.~Girshick, and J.~Sun, ``Faster {R-CNN}: Towards real-time object detection with region proposal networks,'' {\em NeurIPS}, vol.~28, 2015.

\bibitem{Geiger2012CVPR}
A.~Geiger, P.~Lenz, and R.~Urtasun, ``Are we ready for autonomous driving? the {KITTI} vision benchmark suite,'' in {\em CVPR}, pp.~3354--3361, IEEE, 2012.

\bibitem{yu2020bdd100k}
F.~Yu, H.~Chen, X.~Wang, W.~Xian, Y.~Chen, F.~Liu, V.~Madhavan, and T.~Darrell, ``{BDD100K}: A diverse driving dataset for heterogeneous multitask learning,'' in {\em CVPR}, pp.~2636--2645, IEEE, 2020.

\bibitem{salehi2021unified}
M.~Salehi, H.~Mirzaei, D.~Hendrycks, Y.~Li, M.~Rohban, M.~Sabokrou, {\em et~al.}, ``A unified survey on anomaly, novelty, open-set, and out of-distribution detection: Solutions and future challenges,'' {\em Transactions on Machine Learning Research}, no.~234, 2022.

\bibitem{miller2018dropout}
D.~Miller, L.~Nicholson, F.~Dayoub, and N.~S{\"u}nderhauf, ``Dropout sampling for robust object detection in open-set conditions,'' in {\em ICRA}, pp.~3243--3249, IEEE, 2018.

\bibitem{harakeh2020bayesod}
A.~Harakeh, M.~Smart, and S.~L. Waslander, ``{BayesOD}: A {B}ayesian approach for uncertainty estimation in deep object detectors,'' in {\em ICRA}, pp.~87--93, IEEE, 2020.

\bibitem{Kraus_2019}
F.~Kraus and K.~Dietmayer, ``Uncertainty estimation in one-stage object detection,'' in {\em ITSC}, pp.~53--60, IEEE, 2019.

\bibitem{du2022siren}
X.~Du, G.~Gozum, Y.~Ming, and Y.~Li, ``Siren: Shaping representations for detecting out-of-distribution objects,'' {\em NeurIPS}, vol.~35, pp.~20434--20449, 2022.

\bibitem{gasperini2021certainnet}
S.~Gasperini, J.~Haug, M.-A.~N. Mahani, A.~Marcos-Ramiro, N.~Navab, B.~Busam, and F.~Tombari, ``{CertainNet}: Sampling-free uncertainty estimation for object detection,'' {\em RA-L}, vol.~7, no.~2, pp.~698--705, 2021.

\bibitem{duan2019centernet}
K.~Duan, S.~Bai, L.~Xie, H.~Qi, Q.~Huang, and Q.~Tian, ``{CenterNet}: Keypoint triplets for object detection,'' in {\em {ICCV}}, pp.~6569--6578, IEEE, 2019.

\bibitem{EvidentialClassification}
M.~Sensoy, L.~Kaplan, and M.~Kandemir, ``Evidential deep learning to quantify classification uncertainty,'' {\em NeurIPS}, vol.~31, pp.~3183--3193, 2018.

\bibitem{lloyd1982least}
S.~Lloyd, ``Least squares quantization in pcm,'' {\em IEEE Transactions on information theory}, vol.~28, no.~2, pp.~129--137, 1982.

\bibitem{NEURIPS2019_9015}
A.~Paszke, , and {et al.}, ``Pytorch: An imperative style, high-performance deep learning library,'' {\em NeurIPS}, vol.~32, pp.~8024--8035, 2019.

\bibitem{scikit-learn}
F.~Pedregosa and {et al.}, ``Scikit-learn: Machine learning in {P}ython,'' {\em Journal of ML Research}, vol.~12, pp.~2825--2830, 2011.

\bibitem{wu2019detectron2}
Y.~Wu, A.~Kirillov, F.~Massa, W.-Y. Lo, and R.~Girshick, ``Detectron2.'' \url{https://github.com/facebookresearch/detectron2}, 2019.

\bibitem{moore2020fiftyone}
B.~E. Moore and J.~J. Corso, ``Fiftyone,'' {\em GitHub Note: https://github.com/voxel51/fiftyone}, 2020.

\bibitem{cocodataset}
T.-Y. Lin, M.~Maire, S.~Belongie, J.~Hays, P.~Perona, D.~Ramanan, P.~Doll{\'a}r, and C.~L. Zitnick, ``Microsoft {COCO}: Common objects in context,'' in {\em ECCV}, pp.~740--755, Springer, 2014.

\bibitem{Kuznetsova_2020}
A.~Kuznetsova, H.~Rom, N.~Alldrin, J.~Uijlings, I.~Krasin, J.~Pont-Tuset, S.~Kamali, S.~Popov, M.~Malloci, A.~Kolesnikov, T.~Duerig, and V.~Ferrari, ``The open images dataset v4,'' {\em IJCV}, vol.~128, no.~7, pp.~1956--1981, 2020.

\bibitem{Everingham10}
M.~Everingham, L.~Van~Gool, C.~K.~I. Williams, J.~Winn, and A.~Zisserman, ``The {P}ascal visual object classes ({VOC}) challenge,'' {\em IJCV}, vol.~88, no.~2, pp.~303--338, 2010.

\end{thebibliography}

\end{document}